\documentclass[conference]{IEEEtran}
\usepackage[utf8]{inputenc}
\IEEEoverridecommandlockouts

\usepackage{cite}
\usepackage{amsmath,amssymb,amsfonts}
\usepackage{graphicx}
\usepackage{textcomp}
\usepackage{xcolor}
\usepackage{hyperref}
\usepackage{makecell}
\usepackage{subcaption}
\usepackage{float}
\usepackage[pscoord]{eso-pic}

\def\figurename{Fig.}

\makeatletter
\def\ps@IEEEtitlepagestyle{%
  \def\@oddhead{%
    \hfill
    \parbox[b]{.72\textwidth}{\raggedleft\footnotesize
      IEEE 2nd International Conference on Computing, Applications and Systems (COMPAS 2025)\\
      23--24 October 2025, Kushtia, Bangladesh%
    }%
  }%
  \def\@oddfoot{}%
}
\makeatother

\title{Lightweight ML-Based Air Quality Prediction for IoT and Embedded Applications}

\author{\IEEEauthorblockN{Md. Sad Abdullah Sami}
\IEEEauthorblockA{\textit{Department of Electrical and Electronic Engineering} \\
\textit{Bangladesh University of Engineering and Technology}\\
Dhaka, Bangladesh \\ sadabdullahsami47@gmail.com}
\and
\IEEEauthorblockN{Mushfiquzzaman Abid}
\IEEEauthorblockA{\textit{Department of Electrical and Electronic Engineering} \\
\textit{Bangladesh University of Engineering and Technology}\\
Dhaka, Bangladesh \\ abid.mushfiq@gmail.com}
}

\begin{document}

\maketitle


\newcommand{\placetextbox}[3]{
 \setbox0=\hbox{#3}
 \AddToShipoutPictureFG*{ \put(\LenToUnit{#1\paperwidth},\LenToUnit{#2\paperheight}){\vtop{{\null}\makebox[0pt][c]{#3}}}
 }
 }
 \placetextbox{.23}{0.055}{\small{979-8-3315-5525-2/25/\$31.00~\copyright 2025 IEEE}}

\begin{abstract}
This study investigates the effectiveness and efficiency of two variants of the XGBoost regression model namely, the full-capacity and lightweight (tiny) models for predicting concentrations of carbon monoxide (CO) and nitrogen dioxide (NO\textsubscript{2}). Utilizing the AirQualityUCI dataset, collected over one year in an urban area, we comprehensively evaluated the predictive performance based on widely accepted metrics, including Mean Absolute Error (MAE), Root Mean Square Error (RMSE), Mean Bias Error (MBE), and the Coefficient of Determination (R\textsuperscript{2}). Additionally, we assessed resource-focused metrics such as inference time, model size, and peak RAM usage. Results reveal that the full XGBoost model offers superior predictive accuracy across both pollutants. However, the tiny model, while slightly less precise, provides significant computational advantages in terms of drastically reduced inference time and model storage requirements. These findings highlight the feasibility of deploying simplified models in resource-constrained environments without substantially compromising performance, making the tiny XGBoost model particularly suitable for real-time air quality monitoring in IoT and embedded applications. This makes the work relevant for smart city air monitoring applications using edge devices.
\end{abstract}


\section{Introduction}
Air pollution remains a significant concern in many urban areas worldwide. High traffic density, industrial activities, and rapid urban growth contribute to deteriorating air quality. Therefore, monitoring and predicting pollutant levels are vital for public health and environmental protection. Traditionally, air quality forecasting relied on physics-based or chemical dispersion models, which demand detailed environmental data and considerable computational power \cite{b5} \cite{b11}. However, these models often face challenges in adapting to the dynamic, real-time conditions of urban environments. This has prompted a move towards data-driven approaches, especially machine learning (ML) techniques, which can more effectively capture complex nonlinear relationships in environmental data \cite{b6}. 

The AirQualityUCI dataset \cite{b15} is among the most commonly used datasets for air quality modeling. It offers hourly measurements of pollutants collected over a year in an Italian city. Previous research has demonstrated that machine learning models such as Decision Trees (DT), Random Forests (RF), and Support Vector Regressors (SVR) can effectively eXtreme Gradient Boosting (XGBoost) predict pollutant levels from sensor data \cite{b2} \cite{b12}. Among these, ensemble methods like RF and XGBoost have proven to be highly effective \cite{b1} \cite{b14}. However, many of these studies focus primarily on predictive accuracy and often neglect the constraints and limitations related to deploying these models in embedded, real-world edge environments. In such contexts, factors like model size, memory consumption, and interference time are equally crucial \cite{b23} \cite{b24}.

In this study, we introduce a resource-aware framework that employs XGBoost to predict carbon monoxide (CO) and nitrogen dioxide (NO\textsubscript{2}) levels using the AirQualityUCI dataset. We evaluate the models with standard regression metrics such as  Mean Absolute Error (MAE), Root Mean Squared Error (RMSE), Mean Bias Error (MBE), and the coefficient of determination (R²) to measure pollutant concentration accuracy. Additionally, we analyze each model’s computational demands to assess its suitability for deployment. This combined approach of accuracy and efficiency underscores the potential of streamlined gradient boosting models for real-time urban air quality monitoring on limited edge devices. Furthermore, given the relevance of air quality monitoring in IoT-enabled smart cities, we examine whether these models can be efficiently deployed on resource-constrained edge devices.

\section{Literature Review}

ML has transformed air quality forecasting. Traditionally, models relied on detailed physical equations and atmospheric chemistry. Now, newer models use past observation data to predict air quality. This shift benefits urban areas, where pollution patterns are affected by many changing factors \cite{b5} \cite{b11}.
Numerous datasets are available for air quality benchmarking, with the AirQualityUCI dataset being particularly notable for its combination of detail and accessibility. Collected from an Italian city, it provides hourly data on key pollutants, making it a popular choice for evaluating various models \cite{b15}. Researchers have used this dataset to explore different machine learning techniques. Hamza et al. tested several regressors, including RF, SVR, and DT, discovering that ensemble methods often delivered superior accuracy \cite{b2}. Similarly, Kumar et al. found that DT and RF models were especially effective in predicting CO and NO\textsubscript{2} levels \cite{b1}. Some studies have combined ML with time series analysis or deep neural networks, but these approaches often increase complexity and require more computational power \cite{b6} \cite{b14}.  
Among various regressors, XGBoost has proven to be a suitable choice for this project. It offers a good balance of accuracy and speed, especially for structured data. While previous studies have tested different models' predictive abilities on the AirQualityUCI dataset, they often overlooked whether these models could be effectively deployed on embedded devices. Many studies did not evaluate if the models are small and fast enough for such hardware. Recent TinyML research highlights that accuracy alone isn’t sufficient; models also need to be lightweight, responsive, and memory-efficient to function well on edge devices \cite{b23} \cite{b24}. Our work aims to address this gap by building on previous research and shifting the focus toward models that are suitable for embedded deployment. Beyond AirQualityUCI, multi-city open datasets enable broader validation across sensors and locales; for example, the SensEURCity release (2020–2021) provides quality-controlled measurements from dense low-cost networks \cite{b25}. Likewise, the QUANT study offers a long-term, multi-city evaluation dataset (2019–2022) covering 49 commercial sensor systems, supporting robustness checks in real-world settings \cite{b26}.

\section{Methodology}

This work presents a lightweight, deployable ML framework designed to forecast air quality by predicting CO and nitrogen dioxide NO\textsubscript{2} levels. Built around the XGBoost algorithm, the aim is to strike a practical balance between predictive accuracy and computational efficiency, particularly for environments with limited hardware resources. The experiments use the AirQualityUCI dataset, which includes 9358 hourly measurements collected in an Italian city between March 2004 and February 2005. Although the dataset contains information on five pollutants, this study focuses on CO and NO\textsubscript{2}, chosen for their relevance and widespread presence in urban air pollution. Initial preprocessing steps involved discarding non-essential fields (such as timestamp and unnamed columns) and converting the remaining values to a fully numeric format. Any placeholder values indicating missing data were replaced or removed. After cleaning, the dataset was normalized using Min-Max scaling, mapping all values into the [0, 1] range. This helped standardize feature contributions and supported more stable and efficient model training. The data was then split into training and testing sets in a 70:30 ratio to enable consistent evaluation. Two configurations of XGBoost were implemented. The full model uses a higher number of estimators and greater tree depth, while the tiny model operates with reduced complexity to reflect embedded-system constraints. Both were trained and evaluated on the same dataset split to ensure comparability. Beyond prediction accuracy, the study also profiled each model's resource requirements. This included measuring inference time, model file size, and peak RAM usage during prediction. These indicators are particularly important for assessing deployment viability in edge environments such as microcontrollers.

To quantify predictive performance, four widely used statistical metrics were applied: MAE, RMSE, MBE, and R². Together, these metrics offer a comprehensive view of how each model performs in terms of accuracy, error distribution, and explanatory power. Also, These resource-focused metrics are critical for deploying ML models in IoT scenarios, where processing power, memory, and latency must remain minimal for real-time responsiveness. By combining accuracy analysis with resource profiling, the proposed method offers a compelling pathway toward practical, embedded air quality monitoring solutions.

All experiments, including model training and evaluation of resource metrics, were conducted on a desktop computer windows machine using python. While our profiling was conducted on a desktop environment using Python, the lightweight design of the tiny model suggests it could be further optimized for deployment on embedded or edge platforms with constrained resources.

\begin{figure}[!ht]
    \centering
    \includegraphics[width=0.45\textwidth]{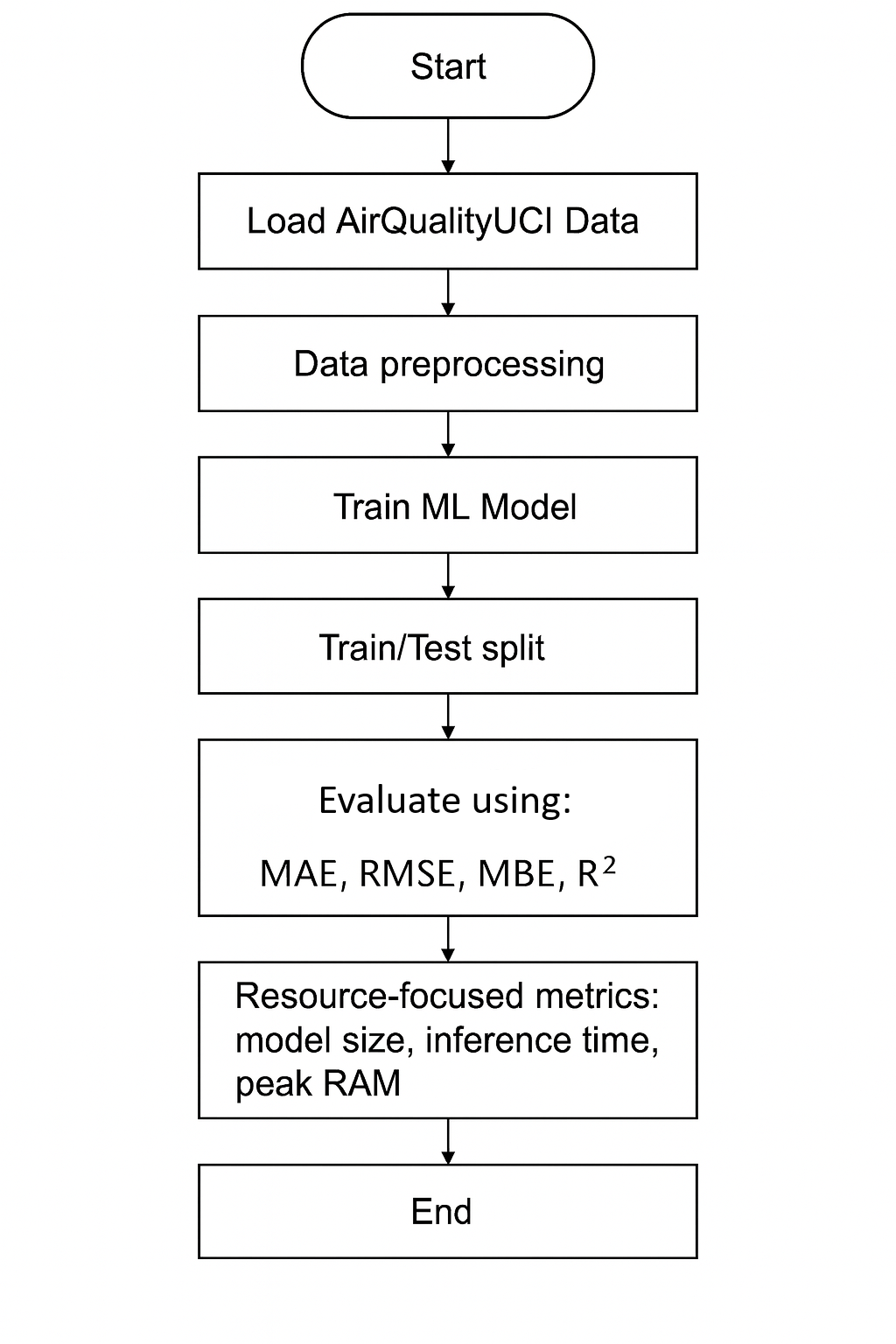}
    \caption{The proposed methodology for pollutant prediction using XGBoost, including preprocessing, training, evaluation and resource-focused profiling.}
    \label{fig:workflow}
\end{figure}

\subsection{Evaluation Metrics}
In order to evaluate the performance of the models for air pollutant concentration prediction, we used four widely used statistical metrics. These are MAE, RMSE, MBE and R\textsuperscript{2}. These metrics offer insights into the magnitude, bias, and variance-explaining capability of each model.

\textbf{Mean Absolute Error (MAE):} MAE means the average absolute difference between predicted and actual values. It is computed as:
\begin{equation}
\text{MAE} = \frac{1}{n} \sum_{i=1}^{n} \left| y_i - \hat{y}_i \right|
\end{equation}
Lower MAE values indicate that predictive performance is better.

\textbf{Root Mean Squared Error (RMSE):} RMSE calculates the square root of the average of squared differences between predicted and actual values. It is computed as:
\begin{equation}
\text{RMSE} = \sqrt{ \frac{1}{n} \sum_{i=1}^{n} (y_i - \hat{y}_i)^2 }
\end{equation}
RMSE punishes the larger errors more than MAE.

\textbf{Mean Bias Error (MBE):} MBE gives the average signed difference between the predicted and the actual values. It is computed as:
\begin{equation}
\text{MBE} = \frac{1}{n} \sum_{i=1}^{n} (\hat{y}_i - y_i)
\end{equation}
 MBE indicates systematic bias. A positive MBE indicates overestimation, while a negative MBE indicates underestimation.

\textbf{Coefficient of Determination (R\textsuperscript{2}):} R\textsuperscript{2} represents the proportion of variance in the dependent variable that is predictable from the independent variables. It is computed as:
\begin{equation}
R^2 = 1 - \frac{\sum_{i=1}^{n}(y_i - \hat{y}_i)^2}{\sum_{i=1}^{n}(y_i - \bar{y})^2}
\end{equation}
An $R^2$ value which is closer to 1 indicates that the model performance is strong.

In addition to predictive performance, we also evaluate the models based on three resource-focused metrics that are essential for embedded deployment:

\textbf{Model Size (kB):} This is the file size of the trained model saved using \texttt{joblib}. It reflects the flash memory requirement and is computed as:
\begin{equation}
\text{Model Size (kB)} = \frac{\text{File Size (bytes)}}{1024}
\end{equation}

\textbf{Inference Time (ms):} Measures how fast the model can make predictions. It is critical for real-time applications and is computed as:
\begin{equation}
\text{Inference Time (ms)} = (t_{\text{end}} - t_{\text{start}}) \times 1000
\end{equation}

\textbf{Peak RAM Usage (MB):} Indicates the maximum memory consumed during model inference. It is measured using the \texttt{memory\_profiler} library:
\begin{equation}
\text{Peak RAM (MB)} = \max(\text{memory\_usage}(f))
\end{equation}

These resource-aware metrics provide additional insights into the deployability of each model in memory-constrained environments.

\subsection{Dataset}

This study utilizes the AirQualityUCI dataset, publicly accessible through the UCI ML Repository. It includes 9,358 hourly averaged observations gathered over a year (March 2004 to February 2005). Data was recorded by an Air Quality Chemical Multisensor Device positioned at street level in a heavily polluted area of an Italian city. The dataset features measurements from five metal oxide gas sensors, alongside ground truth readings from certified analyzers for pollutants such as CO, NO\textsubscript{2}, C\textsubscript{6}H\textsubscript{6}, NMHC, and NO\textsubscript{x}.

\section{Results and Discussion}

\subsection{Comparative Performance}

The predictive performance of the full and tiny XGBoost models for forecasting CO and NO\textsubscript{2} concentrations is summarized in Table~\ref{tab:performance_metrics}. For both pollutants, the full model demonstrates superior accuracy, as indicated by lower MAE and RMSE values and higher R\textsuperscript{2}. The tiny model, although slightly less accurate, achieves commendable predictive capability, especially given its significantly smaller model size and reduced computational complexity.

\begin{table}[ht]
\centering
\caption{Performance comparison of Full and Tiny XGBoost models for CO and NO\textsubscript{2} prediction}
\label{tab:performance_metrics}
\resizebox{\linewidth}{!}{%
\begin{tabular}{|c|c|c|c|c|c|}
\hline
\textbf{Target} & \textbf{Model} & \textbf{MAE} & \textbf{RMSE} & \textbf{MBE} & \textbf{R\textsuperscript{2}} \\
\hline
CO & Full XGBoost & 0.0244 & 0.0381 & -0.000062 & 0.9064 \\
CO & Tiny XGBoost & 0.0344 & 0.0504 & -0.000698 & 0.8356 \\
NO\textsubscript{2} & Full XGBoost & 0.0400 & 0.0546 & -0.002623 & 0.8559 \\
NO\textsubscript{2} & Tiny XGBoost & 0.0558 & 0.0751 & -0.002572 & 0.7266 \\
\hline
\end{tabular}}
\end{table}

Resource usage, shown in Table~\ref{tab:resource_metrics}, reveals substantial advantages of the tiny XGBoost model in terms of model size, inference time, and peak RAM usage. Notably, the tiny model demonstrates significantly faster inference, especially for NO\textsubscript{2} prediction. 

\begin{table}[ht]
\centering
\caption{Resource usage comparison of Full and Tiny XGBoost models for CO and NO\textsubscript{2} prediction}
\label{tab:resource_metrics}
\begin{tabular}{|c|c|c|c|c|}
\hline
\textbf{Target} & \textbf{Model} & \makecell{\textbf{Inference}\\\textbf{Time (ms)}} & \makecell{\textbf{Model}\\\textbf{Size (KB)}} & \makecell{\textbf{Peak}\\\textbf{RAM (MB)}} \\
\hline
CO & Full XGBoost & 2.3902 & 60.9629 & 129.5625 \\
CO & Tiny XGBoost & 2.5173 & 12.7383 & 130.4766 \\
NO\textsubscript{2} & Full XGBoost & 2.0926 & 60.8301 & 130.5313 \\
NO\textsubscript{2} & Tiny XGBoost & 0.9725 & 12.7383 & 130.5703 \\
\hline
\end{tabular}
\end{table}

All resource measurements, including inference time, model size, and peak RAM usage, were obtained on a general-purpose desktop machine using Python. These values are indicative of development-time behavior rather than deployment-level performance on embedded hardware. As such, they may overestimate memory and processing availability compared to typical TinyML environments, which require further model optimization. In particular, peak RAM usage under Python tends to be higher due to interpreter overhead and background processes. 

\begin{figure}[H]
\centering

\begin{subfigure}{0.4\textwidth}
  \centering
  \includegraphics[width=\linewidth]{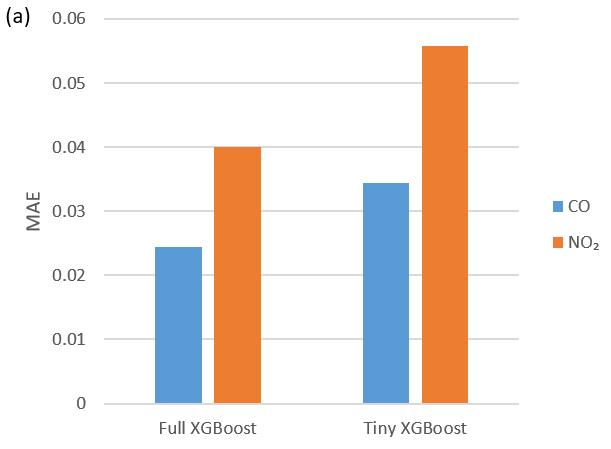}
\end{subfigure}
\hfill
\begin{subfigure}{0.4\textwidth}
  \centering
  \includegraphics[width=\linewidth]{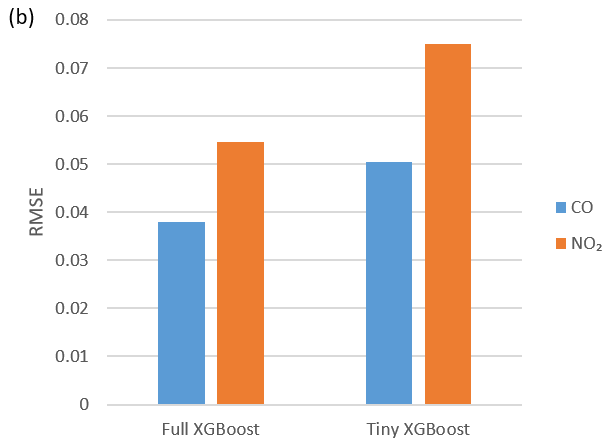}
\end{subfigure}

\vspace{1em}

\begin{subfigure}{0.4\textwidth}
  \centering
  \includegraphics[width=\linewidth]{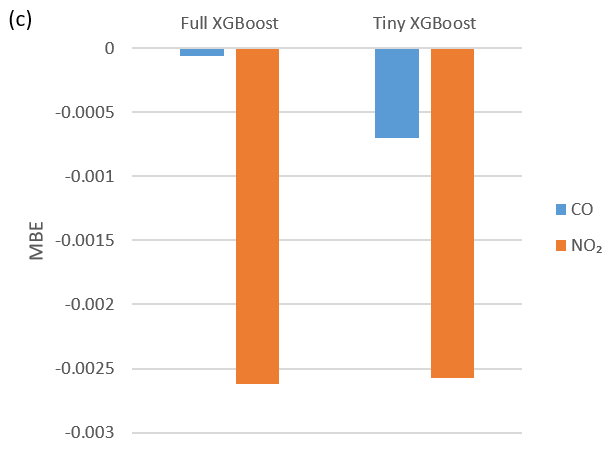}
\end{subfigure}
\hfill
\begin{subfigure}{0.4\textwidth}
  \centering
  \includegraphics[width=\linewidth]{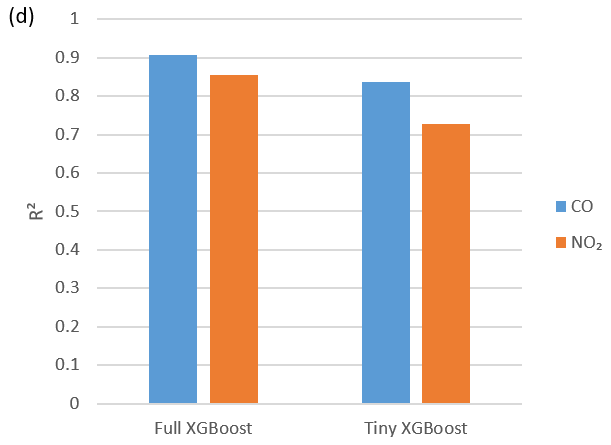}
\end{subfigure}

\caption{Performance comparison of full and tiny XGBoost models on CO and NO\textsubscript{2} concentrations. The subplots show evaluation metrics: (a) MAE, (b) RMSE, (c) MBE, and (d) R\textsuperscript{2}.}
\label{fig:Performance_comparison}
\end{figure}

For deployment on resource-constrained edge devices, especially those in the TinyML domain, additional optimization steps such as quantization, pruning, or model conversion would be necessary. This distinction is crucial when assessing real world feasibility for on-device execution.

\subsection{Analysis of Predictive Metrics}

Figure~\ref{fig:Performance_comparison} provides a visual comparison of the predictive accuracy metrics (MAE, RMSE, MBE, R\textsuperscript{2}) between full and tiny models. The full model consistently outperforms the tiny version across metrics, particularly for NO\textsubscript{2}, indicating its robustness in handling complex nonlinear relationships within the dataset.

\subsection{Analysis of Resource-focused Metrics}

\begin{figure}[H]
\centering

\begin{subfigure}{0.4\textwidth}
  \centering
  \includegraphics[width=\linewidth]{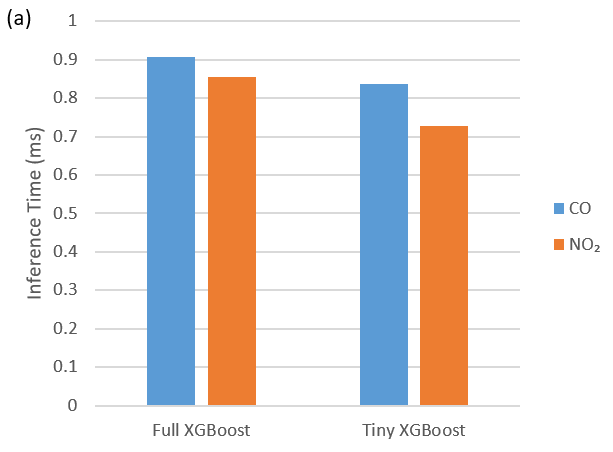}
\end{subfigure}
\hfill
\begin{subfigure}{0.4\textwidth}
  \centering
  \includegraphics[width=\linewidth]{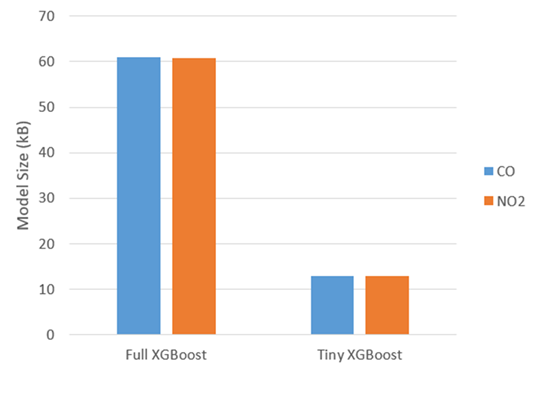}
\end{subfigure}

\vspace{1em}

\begin{subfigure}{0.4\textwidth}
  \centering
  \includegraphics[width=\linewidth]{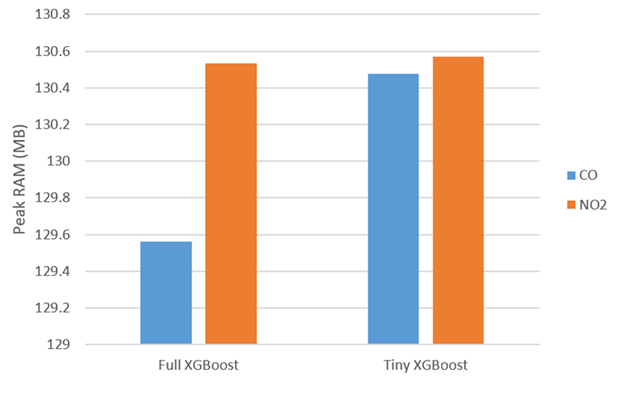}
\end{subfigure}

\caption{Computational resource usage comparison between full and tiny XGBoost models for CO and NO\textsubscript{2} prediction. Metrics include (a) inference time, (b) model size, and (c) peak RAM usage.}
\label{fig:Performance_comparison2}
\end{figure}

The computational efficiency metrics are illustrated in Figure~\ref{fig:Performance_comparison2}. The tiny model shows remarkable reductions in inference time and model size without substantial increases in peak RAM usage. This highlights its suitability for deployment in resource-constrained embedded systems.

\subsection{Trade-off Analysis}

\begin{figure}[htbp]
    \centering
    \includegraphics[width=0.9\linewidth]{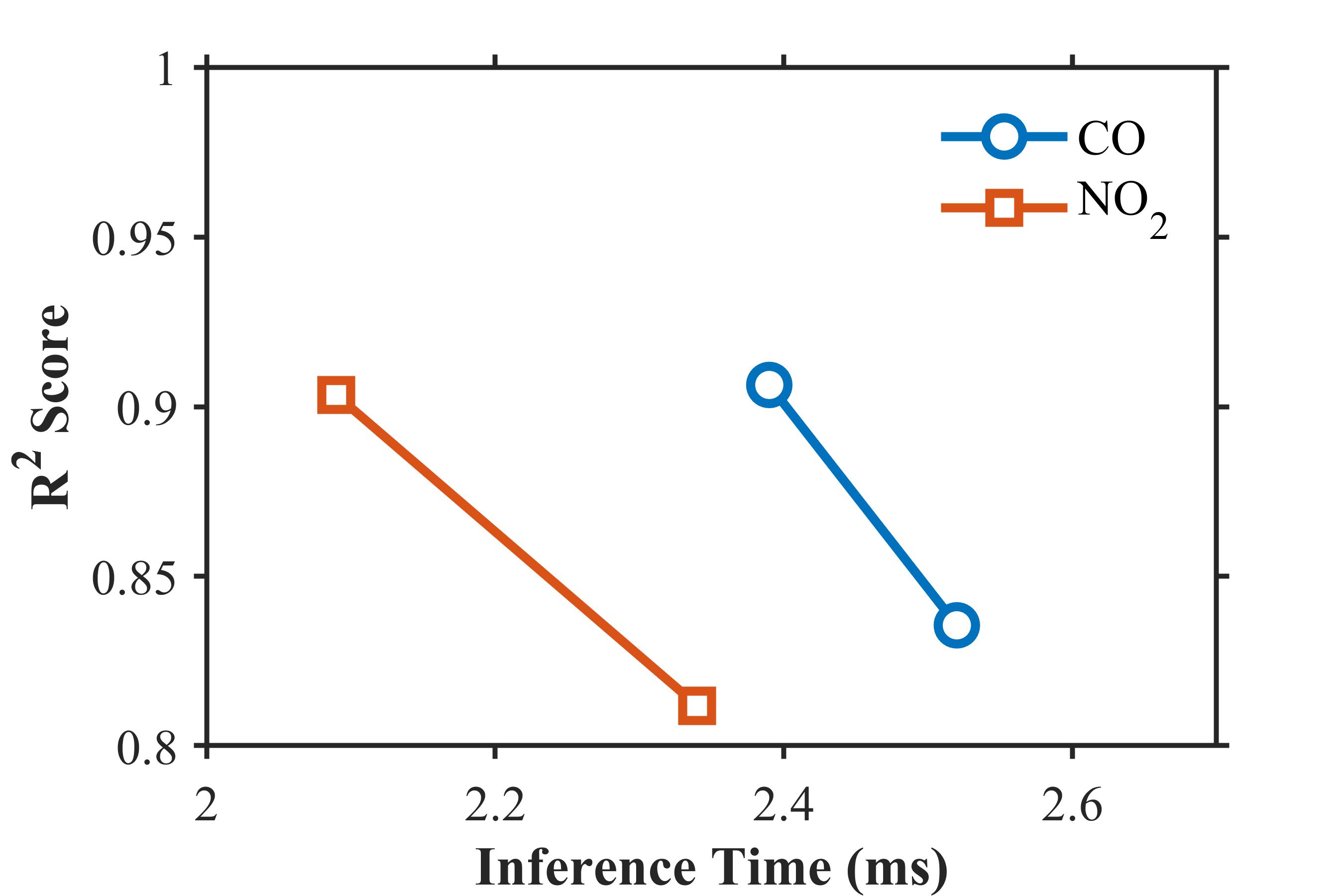}
    \caption{Inference Time vs R\textsuperscript{2} score trade-off curve.}
    \label{fig:inference_tradeoff}
\end{figure}

\begin{figure}[htbp]
    \centering
    \includegraphics[width=0.9\linewidth]{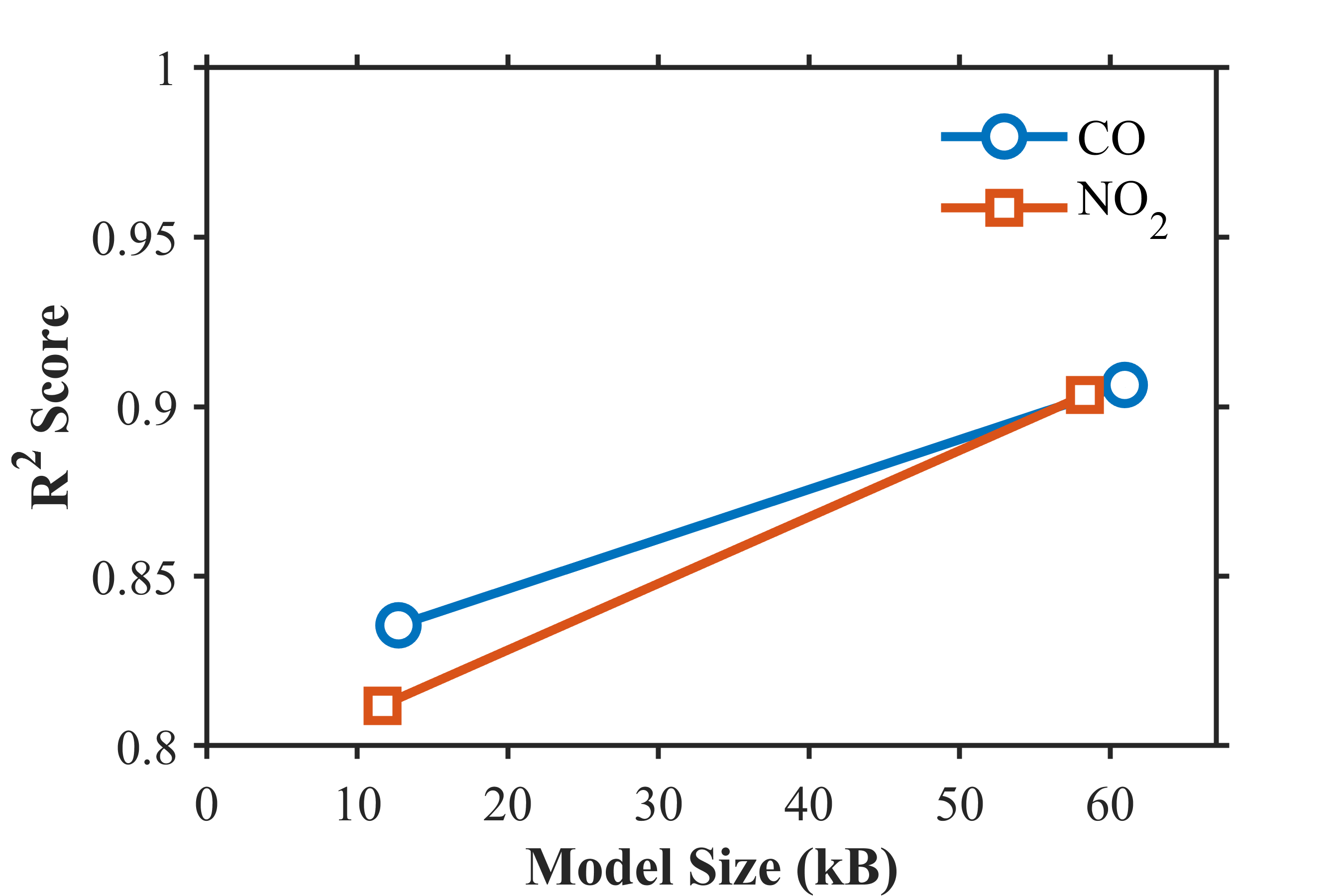}
    \caption{Model Size vs R\textsuperscript{2} score trade-off curve.}
    \label{fig:modelsize_tradeoff}
\end{figure}

While achieving strong predictive performance is often the main focus in many ML applications, deploying models in real-world settings, specially on embedded or resource-limited devices, requires a more balanced approach. Figures~\ref{fig:inference_tradeoff} and Figures~\ref{fig:modelsize_tradeoff} demonstrate this important trade-off by comparing each model's R\textsuperscript{2} score with two key resource-oriented metrics: inference time and model size.

The complete XGBoost model delivers higher accuracy for both pollutants, aligning with expectations due to its deeper trees and greater number of estimators. Nonetheless, this accuracy comes with a significant computational expense. As depicted in Figures~\ref{fig:inference_tradeoff}, the full model takes more than twice as long as the tiny version to perform NO\textsubscript{2} prediction. While this may be acceptable on desktop hardware, it poses challenges on microcontrollers or edge devices where low latency is critical.

The model size trade-off illustrated in Figures~\ref{fig:modelsize_tradeoff} emphasizes an important aspect for deployment on systems with limited memory. The full model requires about 60~kB of storage, which could be challenging for smaller embedded devices with limited flash memory. On the other hand, the tiny model which is simplified in depth and number of estimators, uses roughly 13~kB, making it much easier to deploy without a significant loss in accuracy. Notably, the reduction in R\textsuperscript{2} from the full to the tiny model, although present, stays within a range that is acceptable for many practical forecasting tasks. For instance, a reduction of about 0.07 in R\textsuperscript{2} for CO prediction might be acceptable in scenarios where speed, responsiveness, and energy efficiency are more important than marginal gains in prediction accuracy.

This trade-off is fundamental to TinyML which is selecting models that are sufficiently effective to offer valuable insights while adhering to hardware limitations. In safety-critical or regulated environments, employing the complete model may still be preferable. But for edge applications like smart city monitoring, mobile devices, or IoT endpoints, the tiny XGBoost model offers a good balance. It greatly reduces resource use like storage, computation, and latency, while maintaining most of its predictive ability.

In the end, these plots and metrics show that model choice shouldn't be based on accuracy alone. A clear understanding of the deployment environment is crucial. The results highlight that small architectural tweaks can significantly cut runtime costs, making machine learning more feasible for limited systems without sacrificing reliability.

\section{Limitations}

Several limitations should be considered in interpreting the results of this study. Firstly, the analysis was based on the AirQualityUCI dataset, which covers a one-year span from a single urban locality. Although this dataset is dated (2004--2005), it remains widely used as a benchmark for evaluating air quality prediction models. Its temporal and geographic scope may restrict the generalizability of the models, as pollution patterns can vary significantly across different regions and climatic conditions. Moreover, the relatively short duration does not adequately capture seasonal or long-term variations in pollutant concentrations, which could limit the robustness of predictive insights. 

Another limitation is the offline nature of our experimentation, where models were trained and evaluated on static historical data without validation under dynamic, real-time conditions typical of practical deployment scenarios. In addition, peak RAM usage was profiled in a desktop Python environment, which tends to overestimate memory consumption compared to optimized embedded deployments. Furthermore, the tiny model showed weaker performance for NO\textsubscript{2} (R\textsuperscript{2} = 0.72), which may be less suitable for health-critical applications and indicates the need for further optimization. Lastly, the fixed feature set provided by the dataset might exclude potentially influential variables that could improve predictive accuracy and generalization. 

Future research should focus on validating these models with diverse and extensive datasets, including more recent data from multiple regions and extended periods. Real-time streaming evaluations, comparisons with other lightweight ML methods (e.g., ridge regression, pruned random forests), and the inclusion of additional context-specific features may also help enhance the practical utility and reliability of these models.

\section{Conclusion}

In conclusion, this research systematically evaluated the predictive performance and resource efficiency of full and tiny variants of the XGBoost regression models for forecasting CO and NO\textsubscript{2} concentrations. The comparative analysis clearly demonstrated that while the full model provides optimal predictive accuracy, the tiny model significantly reduces computational demands such as inference latency, memory footprint, and model storage space, making it highly practical for embedded and IoT applications with constrained resources. Our findings underscore the importance of balancing accuracy and computational efficiency when developing predictive models for real-world deployment. Moving forward, future studies should aim to extend this analysis to diverse environmental contexts and implement real-time validation frameworks. Exploring adaptive and context-aware model tuning could further enhance predictive performance and operational practicality, thereby contributing to more robust and versatile air quality monitoring solutions.

\end{document}